%% file: main.tex
 \documentclass[sigconf]{acmart}
 
\usepackage{algorithm}
\usepackage{algorithmic}
\usepackage{xcolor}
\usepackage{multirow}
\usepackage{float}
\usepackage{bm}

\newcommand{\cmark}{\color{red}{$\bm{\checkmark}$}}%

\definecolor{algoGreen}{rgb}{0.11, 0.7, 0.03}
\newcommand{\xmark}{\color{algoGreen}$\bm{\times}$}%

\definecolor{algoGray}{rgb}{0.35, 0.35, 0.35}
\newcommand{\algocomment}[1]{{\color{algoGray} #1}}
\newcommand{\etal}{\textit{et al.}}
\newcommand{\etc}{\textit{etc.}}
\newcommand{\ie}{i.e.}
\usepackage{sidecap}

\AtBeginDocument{%
  \providecommand\BibTeX{{%
    \normalfont B\kern-0.5em{\scshape i\kern-0.25em b}\kern-0.8em\TeX}}}

\setcopyright{acmcopyright}
\copyrightyear{2021} 
\acmYear{2021} 

\acmConference[RecSys '21]{Fifteenth ACM Conference on Recommender Systems}{September 27-October 1, 2021}{Amsterdam, Netherlands}
\acmBooktitle{Fifteenth ACM Conference on Recommender Systems (RecSys '21), September 27-October 1, 2021, Amsterdam, Netherlands}
\acmPrice{15.00}
\acmDOI{10.1145/3460231.3474233}
\acmISBN{978-1-4503-8458-2/21/09}

\begin{document}

\title{Semi-Supervised Visual Representation Learning for Fashion Compatibility}

\author{Ambareesh Revanur}
\authornote{Work done during an internship at Walmart Global Tech Bangalore.}
\email{ambareesh.r@gmail.com}
\affiliation{%
  \institution{Carnegie Mellon University}
  \country{USA}
}

\author{Vijay Kumar}
\affiliation{%
  \institution{Walmart Global Tech Bangalore}
  \country{India}}

\author{Deepthi Sharma}
\affiliation{%
  \institution{Walmart Global Tech Bangalore}
  \country{India}
}

\renewcommand{\shortauthors}{Ambareesh, Vijay and Deepthi}

\begin{abstract}
  \input{section/abstract}
\end{abstract}


\maketitle

\section{Introduction}
\input{section/introduction.tex}

\section{Related Work}
\input{section/related_work.tex}

\section{Our Approach}
\input{section/main.tex}

\section{Datasets}
\input{section/datasets.tex}

\section{Experiments}
\input{section/experiments.tex}

\section{Conclusion}
\input{section/conclusion.tex}

\bibliographystyle{ACM-Reference-Format}
\bibliography{acmart}
 
\end{document}

%% file: section/abstract.tex
We consider the problem of complementary fashion prediction. Existing approaches focus on learning an embedding space where fashion items from different categories that are visually compatible are closer to each other. However, creating such labeled outfits is intensive and also not feasible to generate all possible outfit combinations, especially with large  fashion catalogs. In this work, we propose a semi-supervised learning approach where we leverage large unlabeled fashion corpus to create {\em pseudo} positive and negative outfits on the fly during training. For each labeled outfit in a training batch, we obtain a pseudo-outfit by matching each item in the labeled outfit with unlabeled items. Additionally, we introduce consistency regularization to ensure that representation of the original images and their transformations are consistent to implicitly incorporate colour and other important attributes through self-supervision.  We conduct extensive experiments on Polyvore, Polyvore-D and our newly created large-scale Fashion Outfits datasets, and show that our approach with only a fraction of labeled examples performs on-par with completely supervised methods.

%% file: section/introduction.tex
\begin{figure}[t]
    \centering
    \includegraphics[width=\columnwidth]{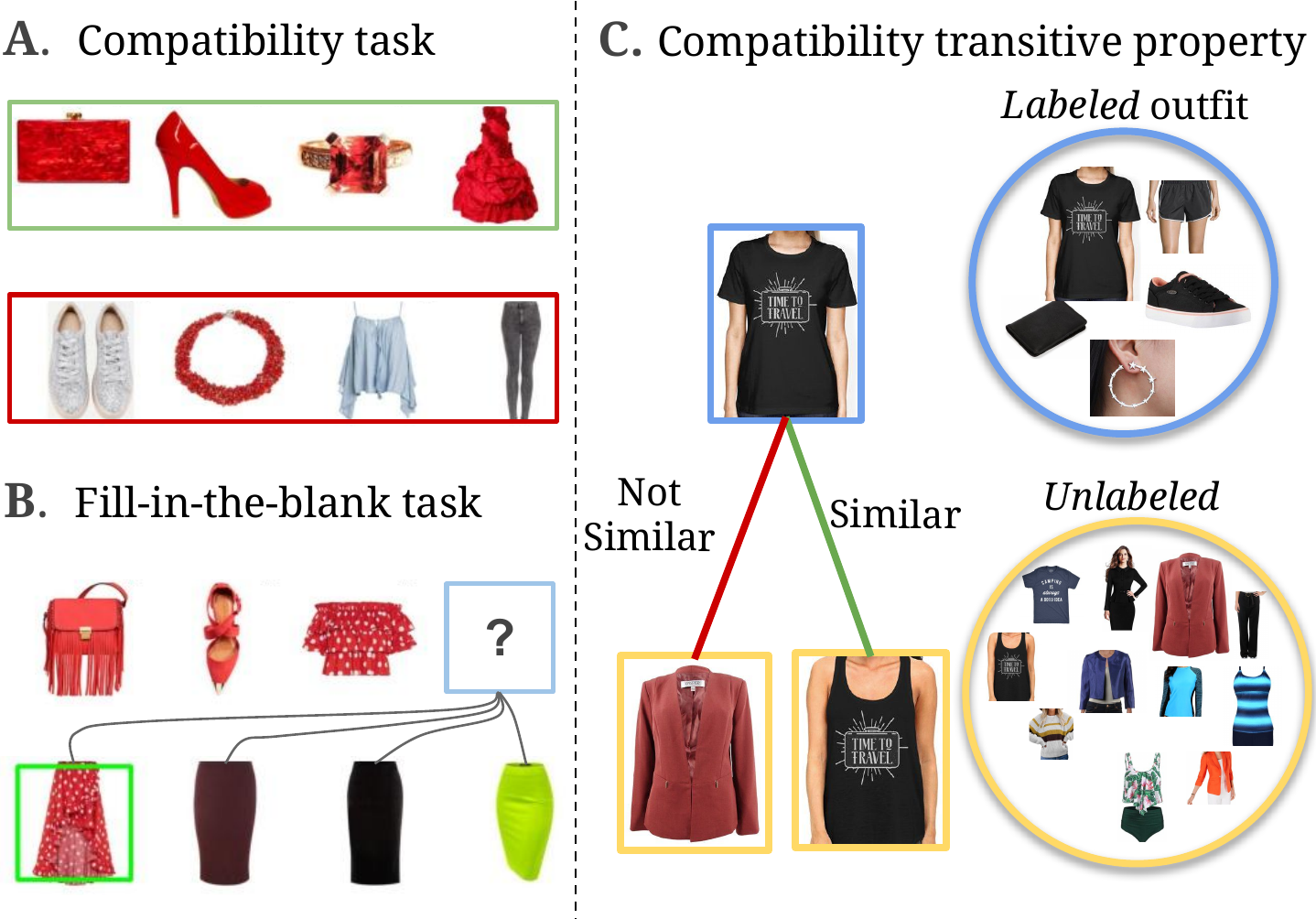}
    \caption{\textbf{(A)} \textbf{Compatibility task.} Compatible (green) and a non-compatible (red) outfit. \textbf{(B)} {\bf Fill in the blank task}. Given an outfit, the objective here is to pick the most compatible item from the given choices. \textbf{(C)} We generate pseudo-outfits based on visual similarity of labeled outfits with unlabeled examples.}
    \label{fig:concept_fig}
\end{figure}

Recent advancements in computer vision have led to their several practical applications in fashion such as similar product recommendation \cite{lu2019learning,cvpr2020fashion,tangseng2020toward,tangseng2020reco}, shop-the-look \cite{hadi2015buy,liu2012street}, virtual try-ons \cite{neuberger2020image,yang2020towards} and 3D avatars \cite{ma2020learning,patel2020tailornet}. In this work, we focus on {\em fashion compatibility} where the objective is to compose matching clothing items that are appealing and complement well, as shown in the Fig.~\ref{fig:concept_fig}{A} and Fig.~\ref{fig:concept_fig}{B}. This could have potential application in online retail industry to recommend complementary products to the user based on their previously purchased product(s). For example, a formal {\em shoe} can be recommended to a customer who purchased a office-wear {\em trouser}.

\vspace{1mm}

What constitutes a good ``complementary'' similarity differs from visual similarity. Learning representations for compatibility requires reasoning about color, shape, product category and other high level attributes to ensure that the items from ``different'' categories are closer. Most existing works \cite{eccv2018learning, iccv2019learning,cvpr2020fashion} achieve this through careful labeling of dataset to identify items that go well together. A simple metric learning approach can then be used to learn an embedding space where the compatible items are brought closer compared to non-compatible ones. However, creating such labeled outfit pairs is expensive, laborious and sometimes require expert knowledge. This also becomes cumbersome and infeasible even for a reasonably large fashion catalog as one needs to compose several possible outfit combinations.
\vspace{1mm}
In this work, we aim to learn powerful representation for compatibility task with very limited labeled outfit data. We propose a semi-supervised learning approach to leverage large amount of unlabeled fashion images that can be easily obtained. The approach is based on data augmentation technique \cite{berthelot2019mixmatch, yun2019cutmix} where the goal is to enhance the training set through techniques namely - {\em pseudo-labeling} and {\em consistency regularization}. Based on the idea that new combinations or associations can be formed from the current associations of the outfit items, we aim to generate {\em pseudo positive} and {\em negative} outfit pairs on-the-fly during each training iteration. For example, consider that if an item A is compatible with another item B and if item C is visually similar to item A, it is possible to then create a new outfit with item pairs B and C. For each image in the positive/negative training pairs, we find the most visually similar example in the unlabeled set and create a new pseudo-outfit pairs as shown in Fig.~\ref{fig:concept_fig}{C}. Thus even with few training outfit collections, it is possible to generate a large stream of pseudo-outfits pairs.

\vspace{1mm}

As image attributes such as colour, shape and texture play a big role for compatibility, we additionally introduce self-supervised consistency regularization \cite{berthelot2019mixmatch} on unlabeled images to explicitly learn those attributes. For instance, we need to disentangle shape information from our representation as items in an outfit are usually of different shape. Similarly, colour can be very informative. We achieve this by applying random transformation on the images and direct the network to produce (dis)similar representations. Note that this is different from explicit attention mechanism employed by current schemes where they train conditional masks to learn a subspace for different attributes such as color and patterns \cite{iccv2019learning,cvpr2020fashion}. 
\vspace{1mm}

We conduct extensive experiments on Polyvore and Polyvore-Disjoint \cite{eccv2018learning} datasets and show that our proposed approach can achieve on-par performance with fully supervised methods even with $5$\% of labeled outfits. In a fully supervised setting, our approach can achieve state-of-the-art performance on compatibility prediction task. Finally, we create a large scale Fashion Outfits dataset consisting of around $700$K outfits with more than $3$M images and demonstrate consistent improvement in performance over fully-supervised baseline. 

\vspace{1mm}

To summarize, we make the following contributions in this paper.

\begin{itemize}
\item We propose a semi-supervised approach for fashion compatibility prediction by learning powerful representation with limited outfit labels.

\item We introduce consistency regularization and pseudo-labeling based data augmentation techniques to learn different attributes and generate pseudo-outfits, completely from the unlabeled fashion images.

\item We demonstrate on-par performance as fully supervised approaches with only a fraction of labeled outfit on Polyvore, Polyvore-D and our newly created datasets.
\end{itemize}

%% file: section/related_work.tex
In this section, we discuss previous works on compatibility prediction and other related areas.

\vspace{1mm} 

\noindent \textbf{Fashion Compatibility} introduced by Han \etal \cite{bilstm} was formulated as a sequential problem and trained a bi-directional recurrent network model that predicts the next compatible item conditioned on previously observed items in the outfit sequence. They also introduced Polyvore dataset in this work. Vasileva \etal \cite{eccv2018learning} train pairwise embedding spaces and employ metric learning to learn representations for fashion compatibility. Representations are learnt for different pairs of categories which is not feasible when the catalog of large category types. Further, they enriched the Polyvore dataset by introducing more challenging evaluation sets by filtering out common items across train and test splits. Tan \etal \cite{iccv2019learning} learn embeddings by relying on a conditional weight branch on two image representations as an attention mechanism for selecting the subspace. Unlike \cite{eccv2018learning}, this work does not require access to the type information during evaluation. 

\vspace{0.5mm}

Another additional ingredient of these works \cite{bilstm,eccv2018learning,iccv2019learning} is the usage of additional meta-data such as text description. They train a text module using word2vec features \cite{mikolov2013distributed} and enforce a visual-semantic embedding (VSE) loss to align text and image representations. In our work, we focus on only visual information. 

\vspace{0.5mm}

More recently, some works \cite{cucurull2019context,yang2020learning,Duannips} have also exploited item-item relationships and trained a graph convolutional network (\textsc{GCN}) by exploiting didactic co-occurrences of items. Further, Lin \etal \cite{cvpr2020fashion} introduce a fashion retrieval problem for images and learn a type-dependent attention mechanism. Most of the existing literature in fashion compatibility focus on the fully supervised paradigm. In contrast, we address a more practical and challenging semi-supervised paradigm without using additional text data. 

\vspace{1mm}
\noindent \textbf{Semi-supervised learning} has seen lot of progress \cite{berthelot2019mixmatch,yun2019cutmix,verma2019interpolation} in the last few years where ever there is a scarcity of labeled data. In the context of deep representation learning, following techniques are widely employed. In consistency regularization \cite{ouali2020semi,cubuk2018autoaugment}, output class predictions are forced to remain unchanged for different augmentations of the input data. This provides regularization to the model to achieve good generalization. In our work, we incorporate consistency regularization to explicitly capture appearance and shape attributes for fashion items. However unlike previous approaches, we minimize the distance between original image and its shape transformation and simultaneously maximize the distance between original image and its colour transformation. This ensures that the model learns shape-invariant features and color variant features.

\vspace{0.75mm}

On the other hand, entropy minimization \cite{kundu2020towards,grandvalet2005semi} aims to obey cluster assumption by forming low density regions between different classes. Pseudo-labeling \cite{lee2013pseudo,cida_eccv,venkat2021classifier} is a type of self-training algorithm that assign hard labels to unlabeled examples based on the maximum prediction probability. In our work, we create pseudo-outfits and augment our training dataset to exploit transitivity of items across different outfits. Generative adversarial network (GAN) \cite{goodfellow2014generative} based approaches have also been proposed for semi-supervised learning, however these are challenging to train \cite{salimans2016improved}. In a related area of few-shot learning, Prototypical networks \cite{snell2017prototypical} exploit the simple inductive bias of a classifier by defining a prototype embedding as the mean of the support set in the latent space. We make use of a simple inductive bias that the nearest neighbour in embedding space should have similar attributes to construct pseudo-outfits. 

\vspace{1mm}
\noindent \textbf{Self-supervised learning} literature is broadly based on solving a pre-defined task that exploits the structure present in the data \cite{moco,simclr,caron2018deep,goyal2019scaling}, or equivariance \cite{noroozi2017representation,kundu2020unsupervised} and invariance \cite{misra2020self} properties of image transformations. For image classification, different pretext tasks such as jigsaw puzzle \cite{noroozi2016unsupervised}, colorization task \cite{deshpande2015learning}, rotation prediction \cite{gidaris2018unsupervised} have been proposed. It has been shown recently that even simple data augmentation methods  \cite{simclr, moco, mocov2} such as colour jittering and gray scale transformations can provide good supervision. These approaches employ contrastive loss to learn consistent representation for an image and its augmentation while treating other images in the batch as negative samples. In our work, we explicitly define positive and negative data augmentations and impose a self-supervised consistency loss on them. However, our regularizer does not require access to labels as required by \cite{moco,simclr} during evaluation.

%% file: section/main.tex
We are interested in learning powerful visual representation for the task of fashion compatibility. This is achieved through metric learning by bringing the embedding of items (from different categories) that go well together in a outfit closer compared to non-compatible items. While the previous works \cite{eccv2018learning,iccv2019learning,cvpr2020fashion} relied mostly on large labeled outfits, in this work, we instead consider only a fraction of labeled outfits and leverage unlabeled fashion items to learn such representations. 

\begin{figure*}
    \centering
    \includegraphics[width=\textwidth]{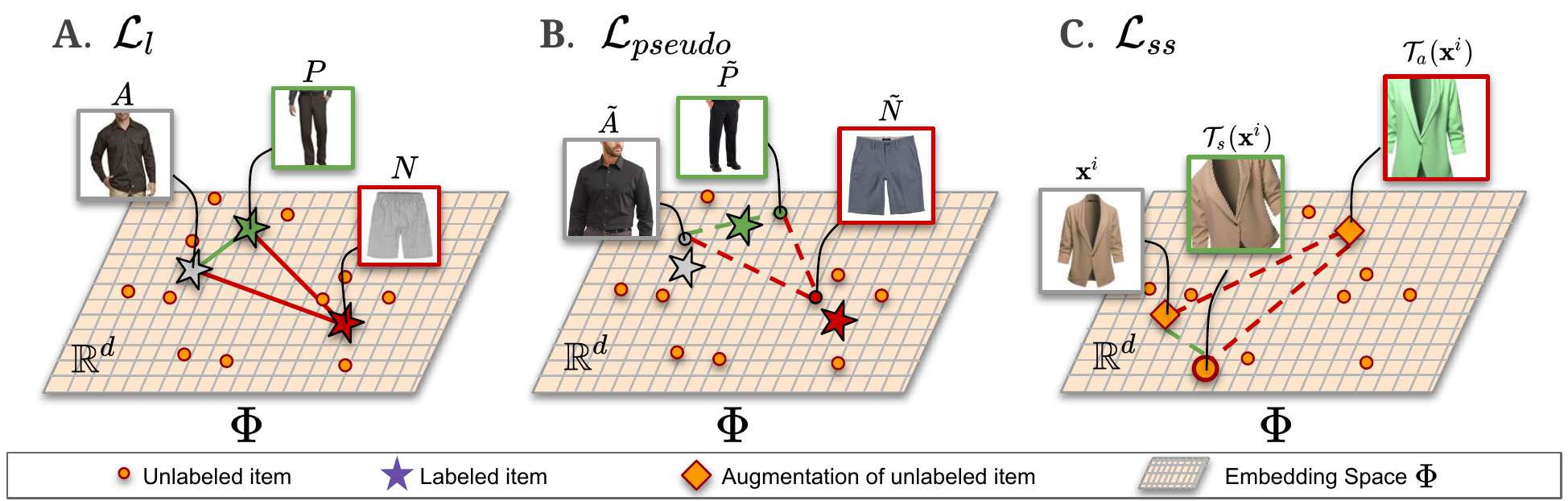}
    \caption{{\bf Overview of our proposed approach}. \textbf{A}. Triplet loss $\mathcal{L}_{l}$ on labeled items: Anchor item $A$, Compatible positive item $P$ and Non-compatible negative item $N$. \textbf{B} Triplet loss $\mathcal{L}_{pseudo}$ on pseudo-labelled items in the unlabeled item batch $\mathbf{b}_U$. The figure shows visually rich low-dimensional embedding space $\Phi$ where we compose nearest pseudo triplets for training. See Sec. \ref{sec:l_pseudo}. \textbf{C}. Triplet losses $\mathcal{L}_{ss}$ on {\em shape} and {\em appearance} transformed images. See Sec. \ref{sec:l_ss}}
    \label{fig:archi}
\end{figure*}

Our proposed approach is a data augmentation technique where {\em pseudo-positive} and {\em negative} triplet pairs (Fig~\ref{fig:archi}(b)) are generated on-the-fly during each training iteration based on visual similarity of individual examples in the triplet using the images in the unlabeled set. Since color and texture play a key role in determining matching outfits \cite{kim2020self}, we incorporate an additional self-supervised loss on the individual fashion items (Fig~\ref{fig:archi}(c)) to implicitly capture those attributes. Note that, this is different from explicit attention mechanism employed by current methods where they learn conditional masks that implicitly learns colour and pattern attributes \cite{iccv2019learning,cvpr2020fashion}.

 We next formulate the problem of semi-supervised fashion compatibility and then describe our proposed data augmentation techniques in greater detail. 
 
\subsection{Formulation}
We are given a dataset $\mathcal{D} = \{\mathcal{D}_{l} \cup \mathcal{D}_{u}\}$ for training our model. $\mathcal{D}_{l}=\{X^1,X^2\dots X^l\}$ corresponds to the labeled set where each  $X^i=\{\mathbf{x}^i_1, \mathbf{x}^i_2,\dots, \mathbf{x}^i_k\}$ is an outfit containing more than one fashion item from different categories. Note that, a fashion item $\mathbf{x}^i_j$ can be a part of multiple outfits. Similarly, $\mathcal{D}_{u}=\{\mathbf{x}^1,\mathbf{x}^2,\dots,\mathbf{x}^u\}$ is the large unlabeled corpus of individual fashion items that are not part of any outfit.
We also denote $\mathcal{X}$ as the set of all fashion images. Our goal is then to learn a mapping function $f_{\theta}:\mathcal{X}\rightarrow\Phi$ that transforms the fashion images into an embedding space where compatible items are closer to each other. In our case, we use deep convolutional neural network (\textsc{CNN}) as $f_{\theta}$ where $\theta$ denotes the parameters of our network.

\subsection{Architecture} Our network architecture is similar to previous approaches \cite{eccv2018learning} that are based on siamese network \cite{siamese2017} with ResNet18 backbone. To train our model, we need a batch of labeled triplets and unlabeled images. Each triplet consists of anchor ($A$), positive ($P$) and negative ($N$) image pairs as shown in the Figure~\ref{fig:archi}. Anchor and positive images are complementary items that are part of the same outfit but from different categories (e.g. {\em shirt} and {\em trouser}) while negative image is chosen randomly from the positive item category. The embeddings $\phi$ are obtained after the fully-connected layer, similar to \cite{eccv2018learning}. We train our network with triplet based margin loss defined as,
\begin{equation}
    \mathbf{max}(0, d(\phi_A, \phi_P) - d(\phi_A, \phi_N) + m) 
\end{equation}
where, $A$, $P$ and $N$ denote anchor, positive and negative images of the triplet, respectively. $d(\cdot, \cdot)$ is the Euclidean distance function and $m$ is the margin.\\

{\bf Labeled dataset:} We directly minimize the above margin loss ($\mathcal{L}_l$) for the triplets sampled from labeled outfits to bring compatible items closer to each other pushing non-compatible items farther away as shown in Fig.~\ref{fig:archi}{A}. However, in the absence of large labeled dataset, we employ consistency regularization and pseudo-labeling losses on the unlabeled images that are described next.

\subsection{Consistency regularization}

\label{sec:l_ss}

Based on the hypothesis that learning discriminative representation for fashion compatibility prediction requires reasoning about important attributes such as colour and texture since matched outfits often have similar attributes. At the same time, it is necessary to disentangle shape information as items from different categories usually have different shapes. This becomes especially crucial when training the ImageNet pre-trained networks that learn discriminative shape information \cite{simclr, moco} for generic image classification. We propose a self-supervised consistency regularization on individual fashion items that enforce these observations on the network. The main difference to existing approaches is that we leverage unlabeled fashion images to explicitly learn such attributes without the need for an attention mechanism \cite{iccv2019learning} or a large labeled outfit dataset \cite{iccv2019learning,cvpr2020fashion}.

\begin{figure}
    \centering
    \includegraphics[scale=0.70]{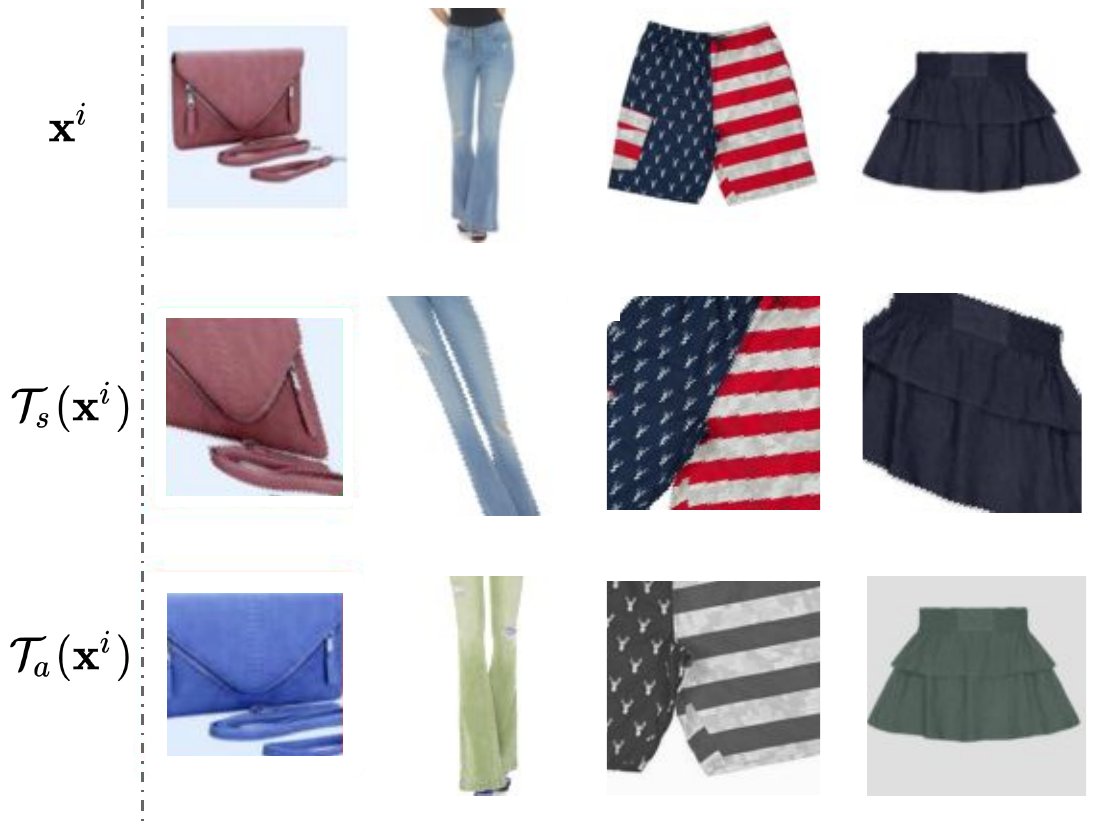}
    \captionof{figure}{Examples showing different \textit{shape} ($\mathcal{T}_s$) and \textit{appearance} transformations ($\mathcal{T}_a$) of the original images (first column). Refer Sec. \ref{sec:l_ss} for details.}
    \label{fig:self_su}
    
\vspace{-3mm}
\end{figure}

We consider entire fashion item set $\mathcal{X}$ that includes both labeled and unlabeled samples without their outfit association. Given each image in the batch, we apply random transformations for shape and appearance, and then measure the discrepancy between the representations of the original and perturbed images. Unlike \cite{berthelot2019mixmatch,yun2019cutmix} that ensures consistent output class distribution for a sample and its transformation, here we enforce the consistency for the representation as network is optimized for distance based loss function. We apply following transformations on the images and visualize a few examples in Fig.~\ref{fig:self_su}.

\begin{itemize}
    \item \textit{Shape transformation ($\mathcal{T}_s$)}: For each image $x^i$ in the mini-batch, we apply random affine transformations such as rotation and shearing. We rotate the image randomly in the range of $\pm5$ degrees and shear the image by utmost $30$ pixels along $x$ and $y$ coordinates. We finally scale the images by a maximum of $1.2$ times and take the center crop of the image. Colour and texture remain unaltered. We represent shape perturbed images as $\mathbf{x}^i_{[s]} = \mathcal{T}_s (\mathbf{x}^i)$. See Fig.~\ref{fig:self_su} for examples.
    
     \item \textit{Appearance transformation ($\mathcal{T}_a$)}: 
    While it is challenging to modify the colour/texture of the items, we resort to simple transformations such as random gray scale, colour jitter and random cropping to obtain the perturbations $\mathbf{x}^i_{[a]}= \mathcal{T}_a (\mathbf{x}^i)$. Such techniques are previously employed in self-supervised works \cite{moco,simclr} for representation learning. See Fig.~\ref{fig:self_su} for examples.
\end{itemize}

Finally, we construct a triplet with original fashion item $x^i$ as anchor, and its shape $\mathbf{x}^i_{[s]}$ and color $\mathbf{x}^i_{[a]}$ perturbed images as positive and negative instances. We then impose a self-supervised loss ($\mathcal{L}_{ss}$) on the augmented triplet $(\mathbf{x}^i, \mathbf{x}^i_{[s]}, \mathbf{x}^i_{[a]})$ using margin triplet loss as shown in Fig.~\ref{fig:archi}{C}.

\subsection{Data Augmentation with Pseudo-Labels}

\label{sec:l_pseudo}
While consistency regularization applied on individual fashion items directs the network what to focus on, we present a labeling strategy that generates {\em pseudo-outfits} by exploiting the knowledge of fashion items in their vicinity. Pseudo-outfits are the synthetically created outfit pairs created on-the-fly during each training iteration. We again leverage unlabeled images to achieve this.

\begin{algorithm}
\caption{Algorithm of our proposed approach}
\label{algo}
\begin{algorithmic}[1]
\small
\STATE \textbf{require:} Labeled $\mathcal{D}_l$ and unlabeled $\mathcal{D}_u$ data, margin $m$, labeled fraction $\alpha$, $\lambda_{pseudo}$ and $\lambda_{ss}$ \algocomment{// $[\cdot]$ is indexing operation} 
\STATE Optimize for model parameters $\theta$ 
\STATE 
\FOR{iteration $i$ in $n_{iters}$} 
	\STATE  \hspace{0.1in} \algocomment{{ // Obtain a batch of fashion items }}
      \STATE \hspace{0.1in} sample $\mathbf{b}_l \sim \mathcal{D}_l$ and $\mathbf{b}_u \sim \mathcal{D}_u$.
       
    \STATE \vspace{1mm} \hspace{0.1in} \algocomment{{// Labeled dataset }}
      \STATE \hspace{0.1in} create ($A$, $P$, $N$) $\sim \mathbf{b}_l$,  type$(P) = $type$(N)$ ~~~ 
      \STATE \hspace{0.1in} $\phi_A, \phi_P, \phi_N \leftarrow f_{\theta}(A), f_{\theta}(P), f_{\theta}(N)$ 
      \STATE \hspace{0.1in} compute $\mathcal{L}_l(\phi_A, \phi_P, \phi_N)$

\STATE  \vspace{1.5mm} \hspace{0.1\in} \algocomment{{// Consistency regularization }}
\STATE \hspace{0.1in} $\mathbf{b}_{[s]} \leftarrow \mathcal{T}_s (\mathbf{b}_u)$, $\mathbf{b}_{[a]} \leftarrow \mathcal{T}_a (\mathbf{b}_u)$  
\STATE \hspace{0.1in} $\tilde{\phi_u}, \tilde{\phi_{[s]}}, \tilde{\phi}_{[a]}\leftarrow f_{\theta}(\mathbf{b}_u), f_{\theta}(\mathbf{b}_{[s]}), f_{\theta}(\mathbf{b}_{[a]})$ 
\STATE \hspace{0.1in} compute $\mathcal{L}_{ss}(\tilde{\phi_u}, \tilde{\phi_{[s]}}, \tilde{\phi}_{[a]})$

\STATE \vspace{1.5mm}\hspace{0.1in} \algocomment{{// Find nearest pseudo-triplet in $\mathbf{b}_u$}}
\STATE \hspace{0.1in} pairwise distance $\text{idx}_A\leftarrow\mathbf{argmin}(d(\phi_A, \tilde{\phi_u}))$. Similarly, compute $\text{idx}_P, \text{idx}_N$ 
\STATE \hspace{0.1in} $\tilde{\phi}_A, \tilde{\phi}_P, \tilde{\phi}_N \leftarrow \tilde{\phi}_u [idx_A], \tilde{\phi}_u [idx_P], \tilde{\phi}_u [idx_N]$ 
\STATE \hspace{0.1in} compute $\mathcal{L}_{pseudo}(\tilde{\phi}_A, \tilde{\phi}_P, \tilde{\phi}_N)$

\STATE \vspace{1.5mm} \hspace{0.1in}\algocomment{{// Minimize the final objective}}
\STATE \hspace{0.1in} update $\theta$ by minimizing $\mathcal{L}$  (Eq. \ref{eq:final_obj})

\ENDFOR
\end{algorithmic}
\end{algorithm}

We draw our motivation from few-shot learning \cite{snell2017prototypical} and argue that nearest neighbour should have similar attributes at a \textit{thematic level} (e.g. different office-wear items to be closer to each other compared to travel-wear items) due to the inductive bias of the network instilled by $\mathcal{L}_{l}$ and $\mathcal{L}_{ss}$.
Thus given a triplet with anchor, positive and negative items from the labeled outfit, we create a pseudo-triplet $(\tilde{A}, \tilde{P}, \tilde{N})$ by finding nearest element for each of these items in the embedding space. As it is computationally infeasible to perform nearest neighbor search on entire unlabeled dataset $\mathcal{D}_U$, we randomly sample sufficiently large mini-batch of unlabeled images, $\mathbf{b}_u$, to generate pseudo-outfits and finally impose margin loss ($\mathcal{L}_{pseudo}$) on the pseudo-triplets $(\tilde{A}, \tilde{P}, \tilde{N})$ as shown in Fig.~\ref{fig:archi}{B}.

\subsection{Loss function}

We formalize our algorithm in Algo. \ref{algo}. We minimize the following objective function to train our model

\begin{equation}
    \mathcal{L}=\mathcal{L}_{l}+\lambda_{ss}\mathcal{L}_{ss}+\lambda_{pseudo}\mathcal{L}_{pseudo}
    \label{eq:final_obj}
\end{equation}
where $\mathcal{L}_{l}$, $\mathcal{L}_{pseudo}$ and $\mathcal{L}_{ss}$ are the triplet margin losses on the labeled, pseudo-outfits and individual instances, respectively. $\lambda_{ss}$ and $\lambda_{pseudo}$ are the hyper-parameters.

%% file: section/datasets.tex
Polyvore outfits \cite{eccv2018learning} and Polyvore disjoint \cite{eccv2018learning} are the two primary datasets used in the literature for evaluating fashion compatibility. We conduct our ablations and comparisons with previous state-of-the-art approaches on these datasets. In addition to these, we baseline our results on a newly created fashion dataset to provide large scale evaluation. We provide the statistics of these datasets in Table~\ref{table:datasets}.

\vspace{2mm}
\noindent \textbf{Polyvore outfits} is a crowd-sourced dataset collected from Polyvore website where users upload fashion photos and organize them into outfits by associating fashion items that go well together. Each outfit consists of product images along with their metadata such as item IDs, product name, fine-grained item type, title and semantic category. There are $12$ semantic categories such as tops, bottoms, outerwear, shoes \etc. There are about a total of $68{,}306$ outfits split into train, valid and test splits as shown in Table \ref{table:datasets}. 
In our work, we demonstrate visual representation learning using only $\alpha\%$ of train split as labeled dataset $\mathcal{D}_L$ and consider rest of the outfit images as unlabeled items $\mathcal{D}_U$. We report our final results on the entire validation and test set as done in previous works.

\vspace{2mm}
\noindent \textbf{Polyvore-Disjoint} dataset is a subset of Polyvore dataset consisting of around $32{,}140$ outfits. It is created by filtering out training outfits that have common items with validation and test outfits. This ensured that train and test outfits have mutually exclusive fashion items for realistic evaluation. Similar to Polyvore dataset, we use only $\alpha\%$ of the training split as $\mathcal{D}_L$ and use all remaining items as $\mathcal{D}_U$. 

\begin{table}

\captionof{table}{Statistics of Polyvore, Polyvore-D and newly created fashion dataset in terms of number of outfits in train, validation and test splits. We also mention overall fashion items in these datasets. Our dataset has $\sim$10 times more outfits than existing datasets.}
\vspace{1.5mm}
\begin{tabular}{l|cccc}
    \hline
        Dataset & Train & Validation & Test & \#items \\
        \hline
        Polyvore & $~53K$ & $10K$ & $~5K$ & $~365K$ \\
        Polyvore-D & $~17K$ & - & $~15K$ & $~175K$ \\
        Fashion Outfits & $~675K$ & $10K$ & $~20K$ & $~3M$ \\
        \hline
    \end{tabular}
    
    \label{table:datasets}
\end{table}

\vspace{2mm}
\noindent \textbf{Fashion outfits} is a proprietary dataset created based on the user purchase transactions on an e-commerce platform. Our collection process makes a reasonable assumption that multiple fashion items from different categories that are purchased by an user in a single session are complementary and go well together. Only those fashion categories that are similar to high-level categories defined in the Polyvore dataset are considered. Based on the user purchase history over a period of time, we retained user sessions (a) with purchases from more than one category, (b) with uniquely purchased items and (c) that do not have multiple items from the same category. We finally apply association mining algorithm \cite{apriori} on these subset to retain highly frequent co-occurring items and remove any noisy transactions.

Overall, the dataset consists of $~705K$ outfits with more than $~3M$ images. We randomly split the dataset into $~675K$ train, $10K$ validation, and $20K$ test outfits. In our experiments, we primarily intend to use the train set as an unlabeled set to demonstrate the efficacy of our proposed approach with unlabeled examples.

\begin{table*}[t]
    \centering
    \caption{Comparison of our approach against previous state-of-the-art methods. {\color{red}Red} color denotes that the configuration is less suitable in low-data regime. We compare our method against certain fully supervised methods that use additional supervision such as text embeddings and type-specific supervision. \underline{Underlined values} indicate best reported fully supervised results on the dataset. We report FITB accuracy and Compatibility AUC. \textit{Higher is better}. See Sec. \ref{sec:results} for more analysis. $\mathcal{D}$ indicate additional unlabeled data from Fashion outfits dataset used for training. Best viewed in color.}
    \vspace{1mm}
    \resizebox{\textwidth}{!}{
    \begin{tabular}{l|ccc|cc|cc}
    
    \hline
    
        \multirow{2}{*}{Method} & \multirow{2}{*}{Labels $\alpha\%$} & \multirow{2}{*}{Uses text labels} & \multirow{2}{*}{Explicit type conditioning} & \multicolumn{2}{c|}{Polyvore-D dataset} &  \multicolumn{2}{c}{Polyvore dataset} \\
        
        \cline{5-8}
        &&&&FITB Acc. & Comp. AUC & FITB Acc.  & Comp. AUC \\
        \hline 
        \multicolumn{8}{c}{\textit{\textbf{Baselines}}} \\
        \hline
        Color attribute & - & \xmark & \xmark & 39.2 & 0.68 & 41.0 & 0.71\\
        Siamese Network \cite{eccv2018learning} & {\color{algoGreen}5\%} & \xmark & \xmark  & 47.0 & 0.77 & 50.3 & 0.78 \\
        
        \hline
        \multicolumn{8}{c}{\textit{\textbf{Fully Supervised Methods}}} \\
        \hline 
        Siamese Network \cite{eccv2018learning} & {\color{red}100\%} & \xmark & \xmark  & 51.8 & 0.81& 52.9 & 0.81 \\
        Bi-LSTM +VSE \cite{bilstm} & {\color{red}100\%} & \cmark & \xmark  & 39.4 & 0.62& 39.7 & 0.65 \\
        CSN T1:1 \cite{csn} & {\color{red}100\%} & \xmark & \cmark & 52.5 & 0.82 & 54.0 & 0.83 \\
        CSN T1:1 + VSE \cite{csn} & {\color{red}100\%} & \cmark & \cmark  & 53.0 & 0.82 & 54.5 & 0.84 \\
        Type-Aware \cite{eccv2018learning} & {\color{red}100\%} & \cmark & \cmark  & 55.2 & 0.84 & 56.2 & 0.86 \\
        SCE-Net (avg) \cite{iccv2019learning}  & {\color{red}100\%} & \cmark & \xmark & 53.6 & 0.82  & 59.1 & 0.88 \\
        CSA-Net \cite{cvpr2020fashion}  & {\color{red}100\%} & \xmark & \cmark & \underline{59.3} & \underline{0.87} & \underline{63.7} & \underline{0.91} \\
        
        \textit{Ours} & {\color{red}100\%} & \xmark & \xmark  & 54.6 & 0.84 & 57.9 & 0.89 \\
  \hline \multicolumn{8}{c}{\textit{\textbf{Ours - Semi Supervised Approach}}} \\
  \hline 
    \textit{Ours} & {\color{algoGreen}5\%} & \xmark & \xmark  & 51.4 & 0.81 & 54.7 & 0.86 \\
    \textit{Ours} + $\mathcal{D}$ & {\color{algoGreen}5\%} & \xmark & \xmark  & \textbf{51.5} & \textbf{0.82} & \textbf{54.9} & \textbf{0.86} \\
    \hline 
    \end{tabular}
    }
    \label{tab:main}
\end{table*}

%% file: section/experiments.tex
\subsection{Implementation Details}
\label{sec:impl}
We use the same implementation procedure as \cite{iccv2019learning, eccv2018learning} and modify ImageNet-pretrained ResNet-18 \cite{resnet} as our backbone. We consider a batch size of $256$ for labeled set where each sample within a batch consists of anchor, positive and negative images. For the unlabeled set, we consider $1024$ individual fashion items. The triplet loss margin $m$ is set to $0.4$. We determine the values of $\lambda_{ss}$ and $\lambda_{pseudo}$ empirically and set them to $0.1$ and $1$, respectively. The network is optimized with Adam \cite{kingma2014adam} optimizer with a learning rate of $5e$-$5$. The network is trained for $10$ epochs and the best results are reported based on a validation set. Our implementation is done in PyTorch \cite{paszke2019pytorch} and trained on Nvidia-V100 GPU machine with $16$GB memory.

\subsection{Evaluation Tasks}
\noindent \textbf{Fill-In-The-Blank (FITB)} is a question and answering task in which model is presented with an incomplete outfit along with four candidate items as possible answers. The task is then to choose a candidate that is most compatible with the given outfit. As done in \cite{eccv2018learning}, we measure the pairwise cosine similarity between the candidate embedding and the average embedding of the outfit and choose the one with highest similarity. The performance is reported as overall accuracy on this task.

\vspace{2mm}
\noindent \textbf{Compatibility prediction} is a binary prediction task where model has to predict whether all the items in a given outfit are compatible or not. We calculate the average pairwise distance between items in the outfit and report the performance as area under the receiver operating curve (\textsc{AUC}).

\subsection{Results}
\label{sec:results}
\noindent \textbf{Baseline methods.} To validate our hypothesis that color plays an important role for compatibility learning, we report our baseline result with color histogram features. As shown in Table \ref{tab:main}, these simple features perform remarkably well and achieve results on par with some deep architectures (Bi-LSTM approach \cite{bilstm}). This motivates us to explicitly encode colour information in our representation from unlabeled images. Another baseline is a siamese network with triplet loss as reported in \cite{eccv2018learning}. Note that these baselines do not have include any other meta data such as text or label information.

\vspace{1mm}

\noindent \textbf{Comparison with state-of-the-art.} We make comparisons with previously reported approaches on Polyvore datasets in Table \ref{tab:main}. All these approaches rely on a fully supervised outfit dataset for representation learning. We also specify whether these approaches rely on any additional metadata information such as category and title. It is clear from the table that our approach achieves on par performance compared to fully supervised methods with only a fraction of labeled outfits and does not require any additional meta-data information.

\begin{figure*}[t]
    \centering
    \includegraphics[width=\textwidth]{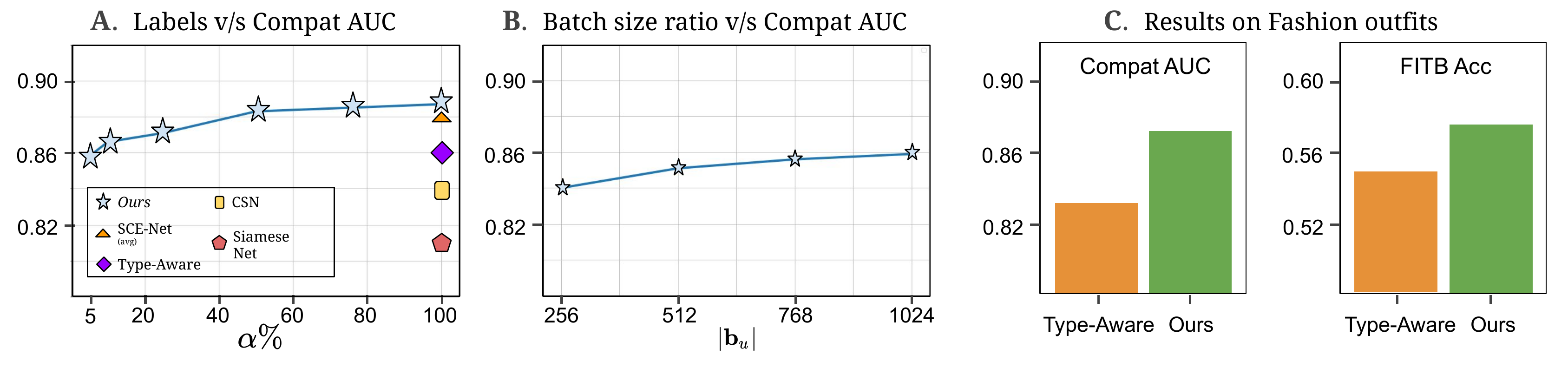}
    \caption{\textbf{A.} Performance of our method with different proportion of training labels ($\alpha\%$) measured by Compatibility AUC on Polyvore dataset. \textbf{B.} Performance of our method with different unlabeled batch size $|\mathbf{b}_u|$ measured by Compatibility AUC on Polyvore dataset. See Sec. \ref{sec:ablations} \textbf{C.} Results of our method and Vasileva \etal  ~\cite{eccv2018learning} on Fashion Outfits dataset. See Sec. \ref{sec:ablations}.}
    \label{fig:additional}
\end{figure*}

\noindent \textbf{Result on Fashion outfits.} We report our large-scale tests on Fashion outfits in Fig. \ref{fig:additional}{ C}. Our method obtains scores of 0.87/57.6 on Compat AUC/FITB tasks while \cite{eccv2018learning} obtains scores of 0.83/55.0. This demonstrates that our method works well even on large  benchmarks. Due to non-availability of code for some methods, we report the result only for \cite{eccv2018learning}.

\begin{figure*}[th!]
    \centering
    \includegraphics[width=\textwidth]{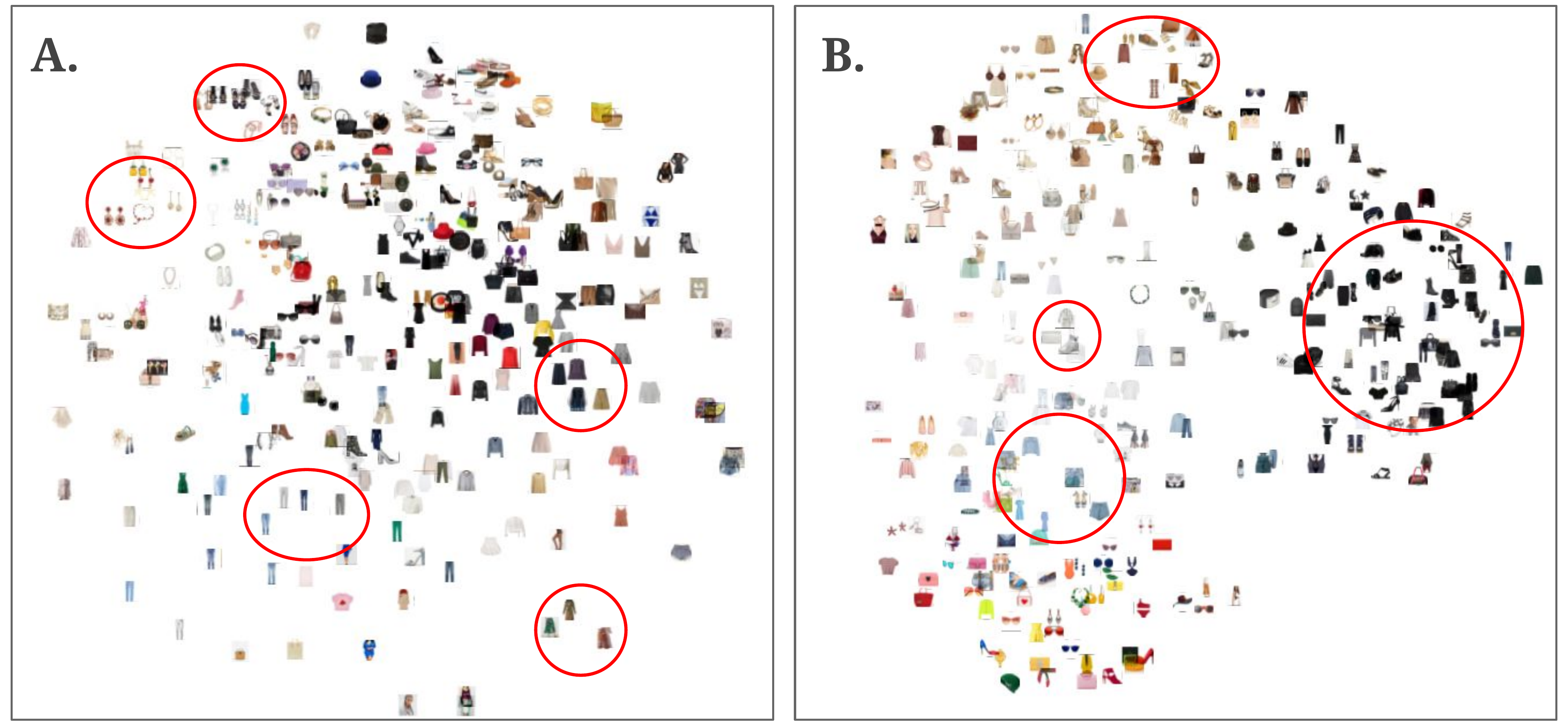}
    \caption{{\bf Visualization of embeddings from} (left) ImageNet pre-trained model. The representations are generally shape-variant as it helps to distinguish different object classes indicating pre-training task bias \cite{simclr,moco}. (right)  Ours ($\alpha=5\%$). The representations show that our embedding space can effectively capture appearance information (such as color). At the same time, items from different categories are closer to each other compared to pre-trained ImageNet model.}
    \label{fig:tsne}
\end{figure*}

\begin{table}[t]
    \centering
    \caption{Ablation studies on Polyvore and Polyvore-D datasets indicating the performance of different components of our model.}
    \vspace{1mm}
    
    \begin{tabular}{l|cc|cc}
    \hline
         \multirow{2}{*}{Model} &  \multicolumn{2}{c|}{Polyvore-D} &  \multicolumn{2}{c}{Polyvore} \\
        \cline{2-5}
        & FITB & Comp.& FITB & Comp. \\
        \hline
        Siamese $\mathcal{L}_{l}$ & 47.0 & 0.77 & 50.3 & 0.78 \\
        $\mathcal{L}_{l} + \lambda_{ss} \mathcal{L}_{ss}$  & 49.2 & 0.79 & 53.7 & 0.82\\
        $\mathcal{L}$ (Eqn~\ref{eq:final_obj}) & 51.4 & 0.81 & 54.7 & 0.86 \\
    \hline 
    \end{tabular}
    \label{tab:ablation}
\end{table}

\subsection{Ablation studies}
\label{sec:ablations}
\noindent \textbf{Consistency regularization and pseudo-labels.} We first conduct our ablation study to understand the effectiveness of $\mathcal{L}_{ss}$ and $\mathcal{L}_{pseudo}$. We include different regularization terms to the baseline siamese model and report the results in Table \ref{tab:ablation}. 
It is evident that both these forms of regularization on unlabeled data have complementary benefits and improve the overall performance significantly achieving results on par with supervised methods.  

\vspace{1mm}
\noindent \textbf{Batch size}. To create good quality pseudo-outfits, we need to have reasonably large unlabeled image batches to ensure better quality matches that share similar attributes for items in the labeled triplet \cite{simclr, lee2013pseudo}. To evaluate this, we conduct an experiment with different unlabeled bath sizes and report the results in Fig \ref{fig:additional}{B}. Results indicate that increasing the batch size consistently improves the performance of our model. However, we could not go beyond a batch size of $1024$ due to \textsc{GPU} memory limitations.

\vspace{1mm}
\noindent \textbf{How many labeled outfits are enough?} Since it is challenging to annotate outfit pairs, an important question we seek to answer is the number of labeled outfits that are sufficient for learning good representations. For this study, we consider a proportion of the Polyvore dataset as labeled and consider remaining outfits as unlabeled in each experiment. As expected, performance improves by increasing the labeled set as shown in Fig \ref{fig:additional}{A}. As depicted in the figure, our semi-supervised method (with $\alpha$~=~$5\%$) outperforms  fully-supervised \textsc{CSN} \cite{csn} and Siamese network. Further, as we increase $\alpha$ to $50\%$, our approach starts to outperform even the fully-supervised approaches that use additional data such as text description \cite{csn,eccv2018learning}. 

Another interesting point to note is that, at full supervision (\ie $\alpha$~=~$100\%$), our approach achieves $0.89$ \textsc{AUC} outperforming many supervised approaches on the compatibility task compared in Table~\ref{tab:main} due to the added consistency regularization supervision.

\subsection{Visualization}

\noindent \textbf{Embedding Space}. We plot t-SNE \cite{tsne} of the visual representation space $\Phi$ for ImageNet and our model as shown in Fig \ref{fig:tsne}. As shown in Fig. \ref{fig:tsne}{A}, ImageNet pre-trained models trained for classification depicts stronger bias for learning shape discriminative features that bringing different category items farther away. This bias hampers the learning of fashion compatibility especially in a semi-supervised setting like ours as discussed in Sec. \ref{sec:l_ss}. In Fig. \ref{fig:tsne}{B}, we portray the t-SNE plot of the representation space learned by our semi-supervised approach ($\alpha$~=~$5\%$). The t-SNE plot demonstrates that the embedding space has learnt strong representations for visual-appearance characterized by the color and texture information of fashion items. Hence, by explicitly disentangling the shape information of the fashion items, our approach overcomes the pre-training task bias. 

\begin{figure*}[t]
    \centering
    \includegraphics[width=\textwidth]{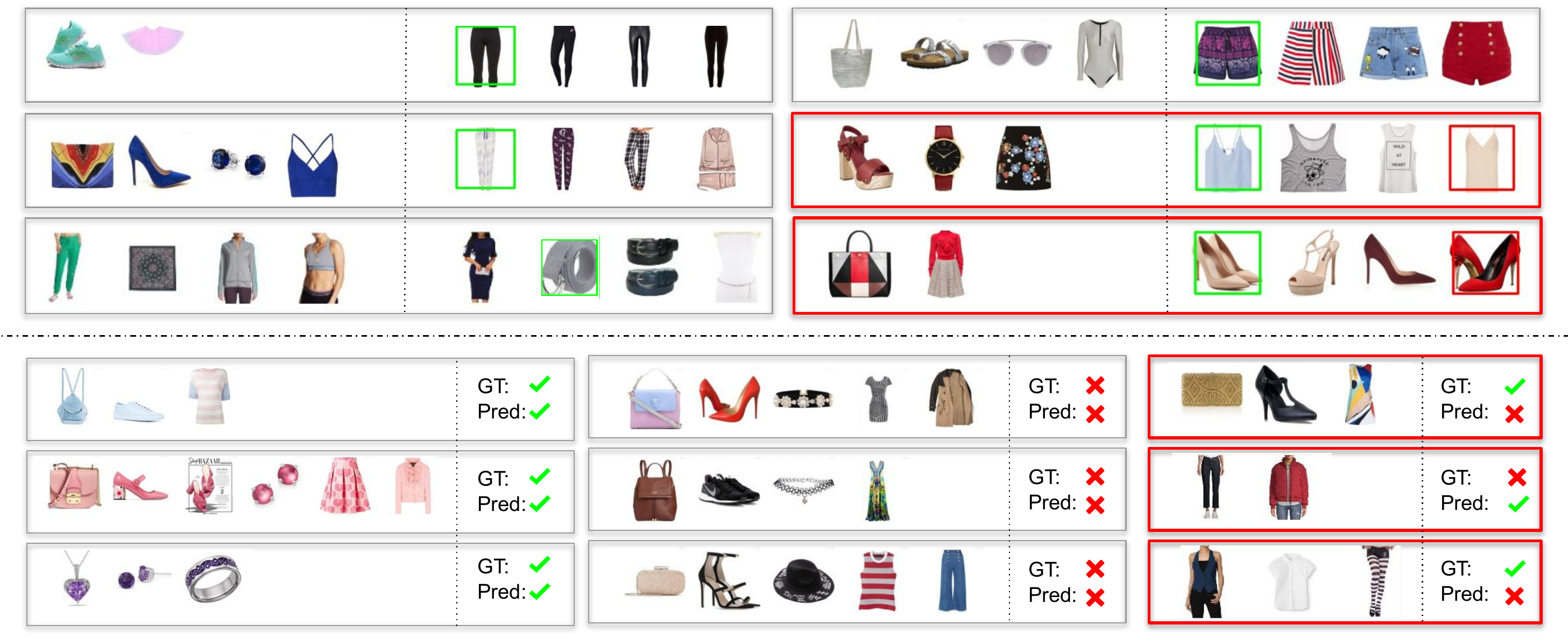}
    \caption{{\bf Qualitative results on Polyvore and Fashion Outfits datasets}. \textbf{Top three rows} show the results on FITB task. Each box contains a query outfit and four candidate choices. Green and red boxes indicates correct and incorrect predictions of our model, respectively. \textbf{Bottom three rows} show the results on compatibility tasks. Some of the failure cases of our model are highlighted with a red box.}
    \label{fig:wm_quali}
\end{figure*}

\vspace{1mm}
\noindent \textbf{Qualitative results}: In Fig. \ref{fig:wm_quali}, we present qualitative results of our approach on the Polyvore and Fashion outfits dataset. The results show that our approach is able to model the concepts of color and texture well. In the Fig. \ref{fig:wm_quali}, we also present some of the failure cases where our approach produces suboptimal predictions for FITB and Compatibility tasks. 
Our model performs suboptimally on outfits that contain items with significantly varying appearance. For example, in the last box of Fig. \ref{fig:wm_quali}, the texture of the \textit{trouser} is very different from that of the \textit{shirt} and the \textit{outerwear}.

%% file: section/conclusion.tex
In this work, we have presented an approach for learning strong visual representations for fashion compatibility using limited labeled training outfit data. We proposed two techniques for leveraging unlabeled data. First, to learn important attributes such as color, we introduced a self-supervision scheme that enforces consistency in the representation of input and its random transformations. While this acts like a data augmentation at instance level, we also proposed pseudo-labeling technique that creates pseudo-labels based on the visual similarity of labeled and unlabeled images. We conducted our experiments on Polyvore, Polyvore-D and newly created Fashion outfits dataset and achieved results on-par with supervised methods. This, however, is achieved with only a fraction of labeled data and without using any meta data such as text description.